\documentclass[10pt,twocolumn,letterpaper]{article}

\usepackage[pagenumbers]{cvpr}

\usepackage{amsfonts,bm}
\usepackage{float}
\usepackage{algorithm}
\usepackage{multirow}
\usepackage{colortbl}

\definecolor{cvprblue}{rgb}{0.21,0.49,0.74}
\definecolor{TableGroup}{HTML}{E8EEF5}
\definecolor{PromptBg}{HTML}{EDF4F1}
\definecolor{QualAccent}{HTML}{3E7771}
\definecolor{TableBest}{HTML}{C23B3B}
\definecolor{TableSecond}{HTML}{2E63B6}


\usepackage[breaklinks,colorlinks,allcolors=cvprblue]{hyperref}

\def\confName{CVPR}
\def\confYear{2027}
\title{TRACE: Trajectory Synchronization for Training-Free \\ Resolution Extension in Video Generation}

\begin{document}


\twocolumn[{%
  \null\vspace{-12pt}
  \begin{center}
    {\Large\textbf{\thetitle}\par}
    \vspace{7pt}
    {\large%
  Libo Zhu\textsuperscript{1}\quad
  Fan Li\textsuperscript{2}\quad
  Zhixin Wang\textsuperscript{2}\quad
  Binglu Wang\textsuperscript{3}\quad
  Weisheng Dong\textsuperscript{4}\quad
  Yulun Zhang\textsuperscript{1}\rlap{\textsuperscript{\textasteriskcentered}}\\[4pt]
  {\normalsize\textsuperscript{1}Shanghai Jiao Tong University} \quad
  {\normalsize\textsuperscript{2}Huawei Technologies Ltd.} \quad \\
  {\normalsize\textsuperscript{3}Northwest Polytechnical University} \quad
  {\normalsize\textsuperscript{4}Xidian University} \quad
\par}
  \end{center}
  \vspace{-5pt}
  \noindent\makebox[\textwidth][c]{%
    \includegraphics[width=\textwidth]{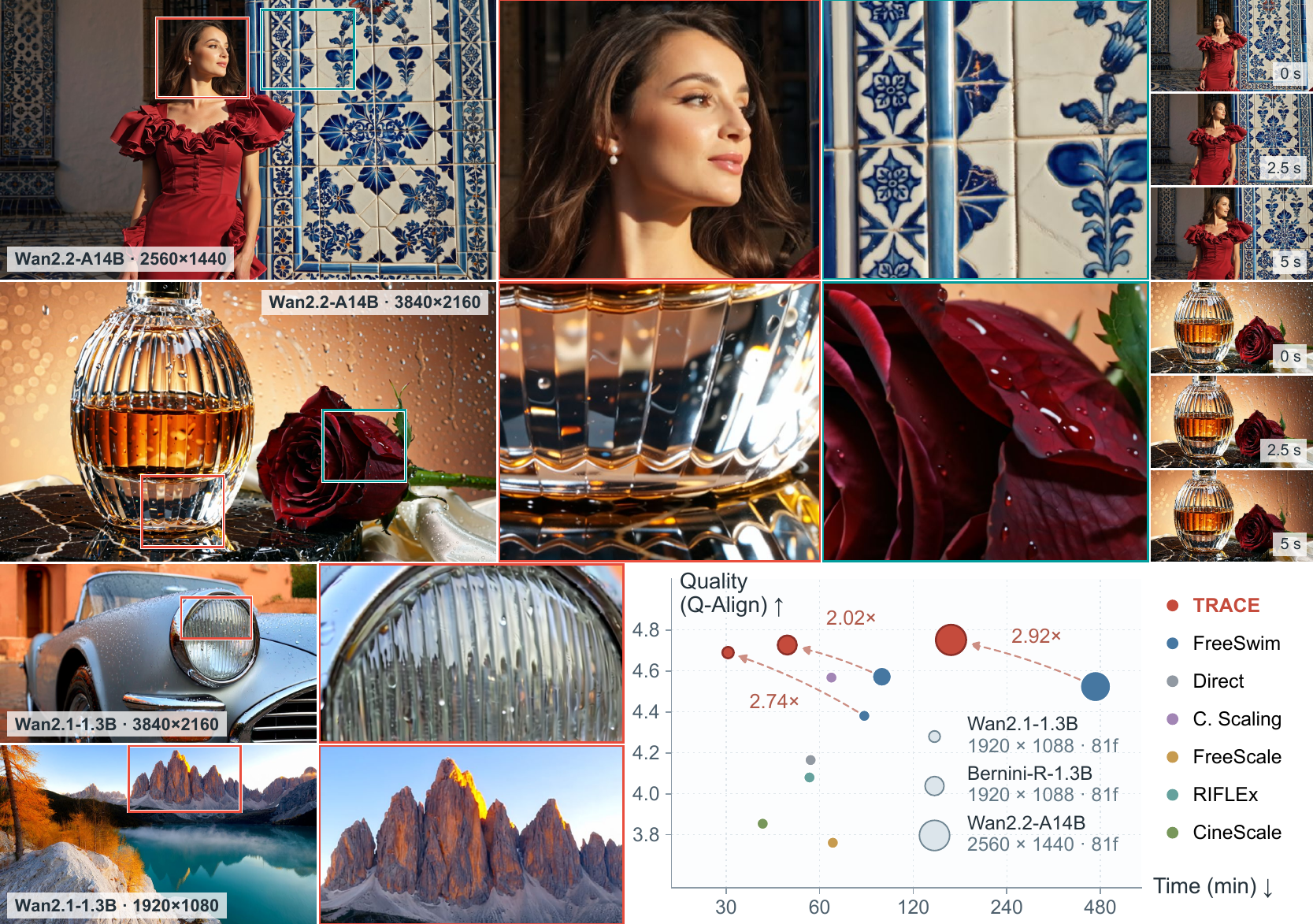}}\par
  \vspace{-7pt}
  \captionof{figure}{\textbf{TRACE} extends existing video generation backbones to higher resolutions while maintaining high-quality generation with rich detail and strong
  temporal consistency. Quantitative comparisons demonstrate its advantages in both
  generation quality and efficiency.}
  \label{fig:teaser}
  \vspace{8pt}
}]
\begingroup
\renewcommand{\thefootnote}{\fnsymbol{footnote}}
\footnotetext[1]{Corresponding author: Yulun Zhang (yulun100@gmail.com).}
\endgroup

\begin{abstract}
  \flushbottom
  \setlength{\parskip}{0pt plus 1pt}
  \predisplaypenalty=10000
  \postdisplaypenalty=10000
  \setlength{\parfillskip}{0pt plus .25\linewidth}
  More text-to-video models are available, but many remain limited to low resolutions. Training them for higher
  resolutions is costly, while existing training-free methods still struggle
  with fine detail, scene structure, and efficiency. We propose \emph{TRACE},
  a training-free method that synchronizes low- and high-resolution generation
  trajectories. During early denoising, TRACE aligns predicted scene content
  at corresponding noise levels. It computes high-resolution updates from
  low-resolution predictions and the current high-resolution state, without
  high-resolution model inference.
  The high-resolution trajectory retains
  fine-grained variation for generating new details. Local window attention then
  refines fine details in a shared high-resolution latent state.
  TRACE checks suspected duplicate subjects against the corresponding
  low-resolution layout and corrects selected regions and their solver history
  while preserving detail
  elsewhere. Experiments across all kinds of video generation backbones and resolutions show improved perceptual quality and
  text--video alignment over FreeSwim, a state-of-the-art training-free method,
  with \mbox{2.0--2.9$\times$} end-to-end speedups. Our code and models will be released at \url{https://github.com/libozhu03/TRACE}.
\par\vspace{-12pt}
\end{abstract}

\twocolumn[{%
      \begin{minipage}{\textwidth}
        \centering
        \includegraphics[width=.9\linewidth]{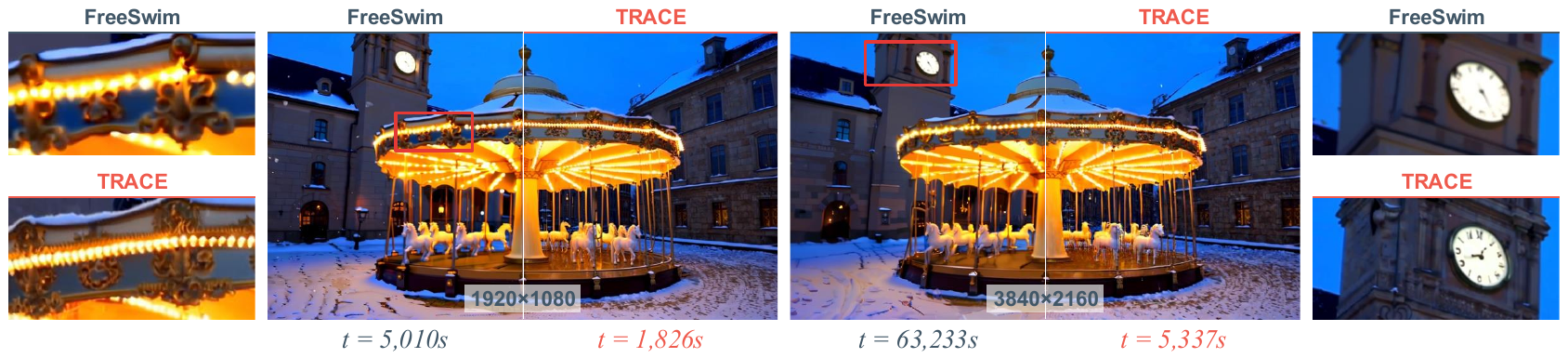}\par
        \vspace{-8.53581pt}
        \captionof{figure}{Visual comparison on \textbf{Wan2.1-T2V-1.3B} at
          $1920\!\times\!1080$ and $3840\!\times\!2160$.
          Central panels show FreeSwim (left) and \textbf{TRACE} (right).
          Outer columns show corresponding detail crops, with FreeSwim above and TRACE below.}%
\label{fig:intro-visual}
      \end{minipage}
      \vspace{12pt}
    }]

\begingroup
\clubpenalty=10000
\widowpenalty=10000
\flushbottom
  \setlength{\parskip}{0pt plus 1pt}
  \interlinepenalty=10000
  \predisplaypenalty=10000
  \postdisplaypenalty=10000
  \setlength{\parfillskip}{0pt plus .25\linewidth}
\section{Introduction}
\label{sec:intro}

Despite recent advances in text-to-video generation~\citep{yang2025cogvideox,kong2024hunyuanvideo,wanteam2025wan},
many models remain limited to relatively low output resolutions.
Extending them to higher resolutions through training is
computationally expensive, while directly running inference on larger grids
can introduce blurred details and structural errors.
It limits the practical use of existing models for
high-quality, high-resolution video generation.

Existing training-free methods address different aspects of high-resolution generation. CineScale~\citep{qiu2025cinescale} adapts positional encodings to improve detail and
scene structure without fine-tuning.
FreeSwim~\citep{wu2026freeswim} combines window attention for local detail with a global branch
for scene structure, using caching to reduce computation.
S$^2$Flow~\citep{wang2026s2flow} uses low-resolution trajectories to construct sparse attention
windows and improve denoising through a second-order
solver. However, existing methods still suffer from semantic drift,
overly smooth visual details, and high inference costs.

S$^2$Flow~\citep{wang2026s2flow} and HiFlow~\citep{bu2025hiflow} use low-resolution information to guide high-resolution predictions.
Despite this guidance, early sampling may still be affected by structural
errors in high-resolution predictions. We observe that early denoising primarily establishes semantic structure:
low-resolution predictions already capture subjects, layout, and motion,
whereas global high-resolution inference can distort this structure.
This motivates constructing synchronized early updates
directly from low-resolution predictions and the current high-resolution
state while retaining fine-grained variation for later detail generation.
These early updates require no high-resolution model inferences.

FreeSwim~\citep{wu2026freeswim} upsamples and re-noises the final
low-resolution result to initialize sampling. Although
this matches the noise level, it does not recover a state consistent
with the corresponding stage of the low-resolution trajectory. Bridging
these mismatched trajectories can introduce structural errors and
semantic drift. We instead construct the early high-resolution trajectory
from intermediate low-resolution predictions, maintaining semantic and
structural alignment across corresponding stages. Refinement
starts from a trajectory consistent with low-resolution generation,
helping reduce local structural errors and semantic drift.

\pagebreak
However, aligning high-resolution sampling with low-resolution trajectories remains challenging.
\textbf{(i)} Low- and high-resolution states differ in representation
and noise, so resizing predictions cannot align their trajectories.
We must account for these differences while preserving variation for new detail.
\textbf{(ii)} Later high-resolution inference can still damage scene
structure. Global attention may blur structural details, while window
attention can produce repeated patterns because each window sees only part of the scene. We must
identify and correct these errors without removing useful new details
in other parts of the generated video.

We introduce \emph{TRACE} (\textbf{TRA}jectory Syn\textbf{C}hronization for Training-Free
Resolution \textbf{E}xtension in Video Generation).
\textbf{Cross-scale trajectory synchronization (CSTS)} starts with shared
coarse noise across resolutions and adds independent fine-grained noise at
high resolution. It calibrates upsampled low-resolution predictions with a
one-time VAE calibration. At each early step, we compute the velocity
from the calibrated prediction and the current high-resolution
noisy state. The solver then advances the trajectory without high-resolution
model inferences.

During subsequent high-resolution denoising, TRACE uses \textbf{window
  attention} on a shared latent grid to reduce computation. Local windows
can still duplicate subjects because of their limited spatial context.
\textbf{Trajectory-guided structural correction (TGSC)} identifies
suspected duplicates from attention maps and checks them against the
low-resolution layout at the corresponding noise level. It corrects
selected regions and their solver history while leaving the rest of the
high-resolution state unchanged.

\begin{figure*}[t]
  \centering
  \includegraphics[width=\linewidth]{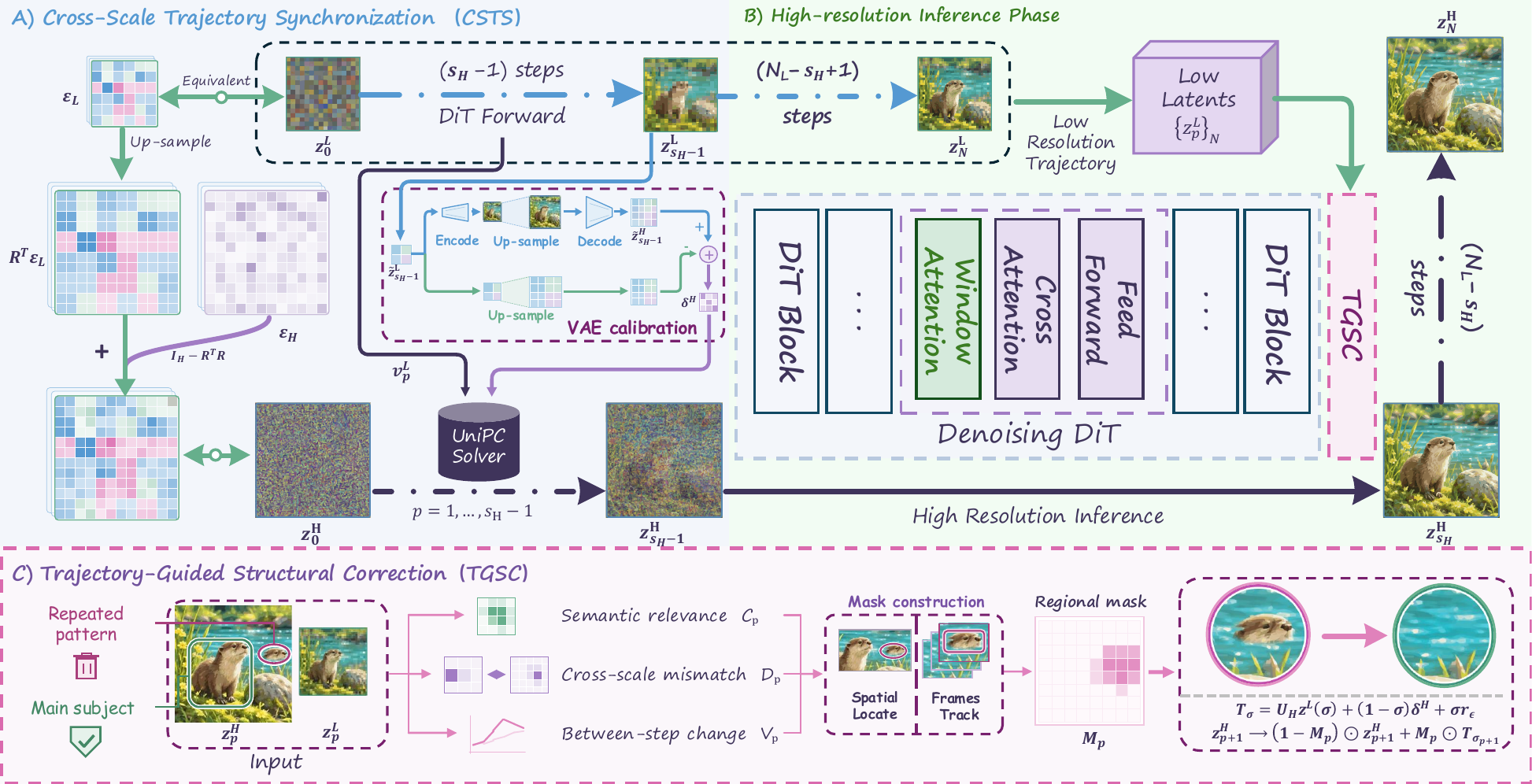}
  \vspace{-17.07164pt}
  \caption{Overview of \textbf{\emph{TRACE}}. The low-resolution trajectory (\textbf{\emph{Low}})
    provides scene references for the high-resolution trajectory (\textbf{\emph{High}}).
    (A) \textbf{\emph{CSTS}} couples their initial noise and computes early High updates
    from Low's predictions, avoiding high-resolution model inferences.
    (B) Window attention computes high-resolution model predictions.
    (C) \textbf{\emph{TGSC}} uses semantic relevance, cross-scale mismatch and between-step change to select suspected duplicated subjects.
    It corrects those latent regions and the corresponding solver history.}
  \label{fig:method-overview}
\end{figure*}

We evaluate TRACE across widely used video generation backbones at commonly
used high resolutions. TRACE improves perceptual quality and text--video
alignment over FreeSwim~\citep{wu2026freeswim}, a state-of-the-art training-free method.
It also achieves \textbf{2.0--2.9$\times$ end-to-end speedups}.
Our contributions are summarized as follows:
\begin{itemize}
  \item We introduce CSTS to synchronize early denoising across
        resolutions without high-resolution model inferences.
  \item We develop TGSC to correct duplicate subjects while preserving
        detail outside the corrected regions.
  \item Our experiments show better generation quality and
        \mbox{2.0--2.9$\times$} speedups over FreeSwim~\citep{wu2026freeswim}.
\end{itemize}
\endgroup

\clubpenalty=10000
\widowpenalty=10000
\displaywidowpenalty=10000
\flushbottom
  \setlength{\parskip}{0pt plus 1pt}
  \interlinepenalty=10000
  \predisplaypenalty=10000
  \postdisplaypenalty=10000
  \setlength{\parfillskip}{0pt plus .25\linewidth}
\flushbottom
\section{Related Work}
\label{sec:related}
\noindent\textbf{Video generation models.}
Early video diffusion operated in pixel space~\citep{ho2022video}.
Cascaded generation~\citep{ho2022imagen} and latent compression~\citep{blattmann2023align}
then enabled generation at larger scales. To model motion across an entire clip,
Lumiere~\citep{bartal2024lumiere} introduced a space-time U-Net.
Later models shifted toward transformers: CogVideoX~\citep{yang2025cogvideox}
couples a 3D VAE with text-video fusion, while
HunyuanVideo~\citep{kong2024hunyuanvideo} and Wan~\citep{wanteam2025wan}
scale transformer backbones and training data to improve visual quality,
motion modeling, and alignment with text descriptions.

\noindent\textbf{Generation acceleration.}
Caching~\citep{zhao2025pab,liu2025teacache,kahatapitiya2025adacache,ma2025magcache}
and sparse attention~\citep{zhang2025sta,gu2026blade} reduce per-step computation.
Distillation~\citep{li2024t2vturbo,yin2025causvid,huang2025selfforcing,zhang2025turbodiffusion}
instead reduces sampling steps, while xDiT~\citep{fang2024xdit} parallelizes inference.
Bottleneck Sampling~\citep{tian2025bottleneck} further lowers intermediate sampling resolutions.

\noindent\textbf{Resolution extension.}
Early extensions adapt receptive fields~\citep{he2024scalecrafter} and feature sizes~\citep{zhang2024hidiffusion},
later targeting video positional embeddings~\citep{qiu2025cinescale}.
Coarse-to-fine sampling instead re-noises low-resolution outputs~\citep{du2024demofusion,guo2024cheapscaling,kim2024diffusehigh}
or fuses frequency components~\citep{qiu2025freescale}.
To control structural errors, AccDiffusion~\citep{lin2024accdiffusion} uses patch-specific prompts,
and FreeSwim~\citep{wu2026freeswim} balances window and global attention.
Beyond final outputs, HiFlow~\citep{bu2025hiflow} aligns flow predictions,
JoLT~\citep{koroglu2026jolt} couples latent trajectories,
FrescoDiffusion~\citep{casellesdupre2026fresco} regularizes tiles, and
S$^2$Flow~\citep{wang2026s2flow} guides sparse windows and solver updates with intermediate low-resolution predictions.

\pagebreak
\noindent\textbf{Video super-resolution.}
Early VSR emphasized flow alignment in BasicVSR++~\citep{chan2022basicvsrpp}
and recurrent attention in RVRT~\citep{liang2022rvrt}.
For richer details, Upscale-A-Video~\citep{zhou2024upscaleavideo} and
VideoGigaGAN~\citep{xu2025videogigagan} introduced diffusion and GAN priors with temporal modeling.
Scaling these priors, PatchVSR~\citep{du2025patchvsr} combines patch and global context,
while FlashVSR~\citep{zhuang2026flashvsr} distills one-step streaming models.

\vspace{3pt}
\section{Methods}
\vspace{4pt}
\label{sec:methods}

\begin{figure*}[t]
  \centering
  \setlength{\abovecaptionskip}{3pt}
  \setlength{\belowcaptionskip}{0pt}
  \noindent\hbox to\linewidth{%
    \includegraphics[width=.245\linewidth]{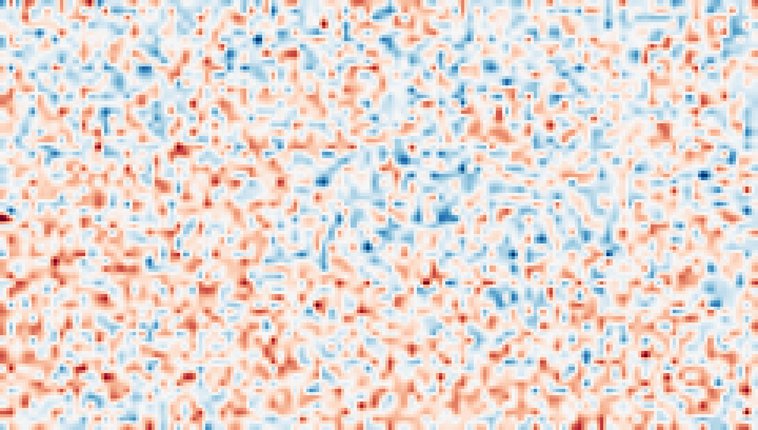}\hfill
    \includegraphics[width=.245\linewidth]{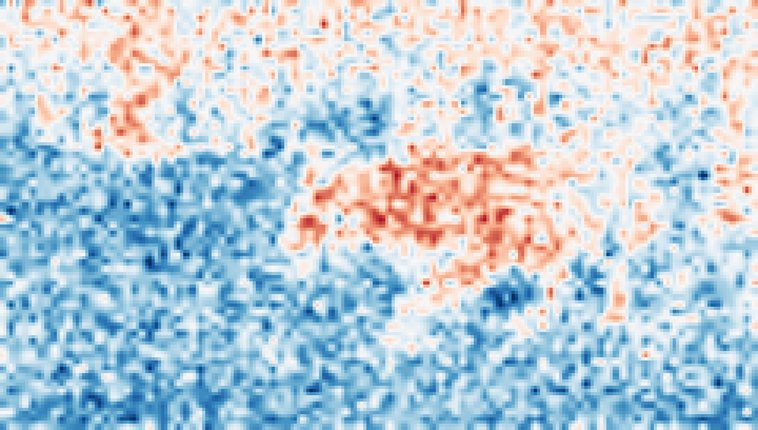}\hfill
    \includegraphics[width=.245\linewidth]{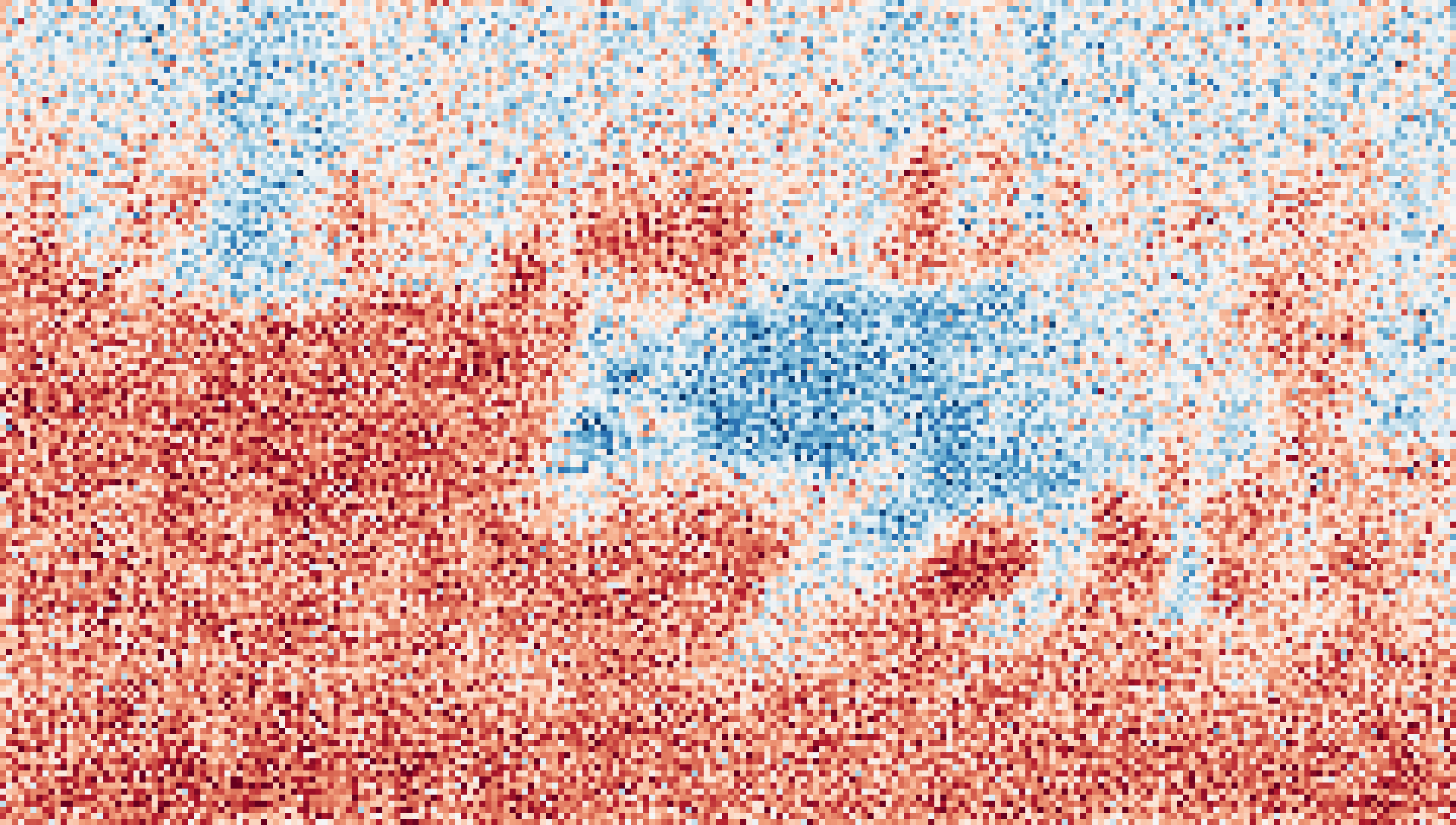}\hfill
    \includegraphics[width=.245\linewidth]{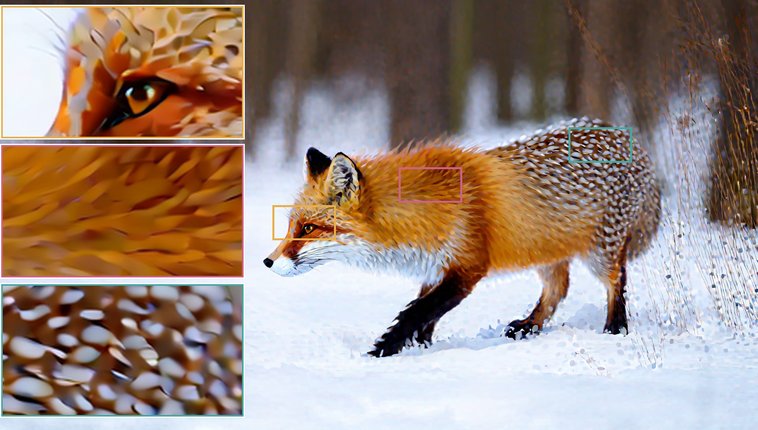}}%
  \par\nointerlineskip\vspace{1.5pt}
  \noindent\hbox to\linewidth{%
    \makebox[.245\linewidth]{\scriptsize\strut (a) Upsampled Low state}\hfill
    \makebox[.245\linewidth]{\scriptsize\strut (b) Upsampled velocity}\hfill
    \makebox[.245\linewidth]{\scriptsize\strut (c) Unaligned Clean prediction}\hfill
    \makebox[.245\linewidth]{\scriptsize\strut (d) Unaligned result}}%
  \par
  \nointerlineskip\vspace{3pt}
  \noindent\hbox to\linewidth{%
    \includegraphics[width=.245\linewidth]{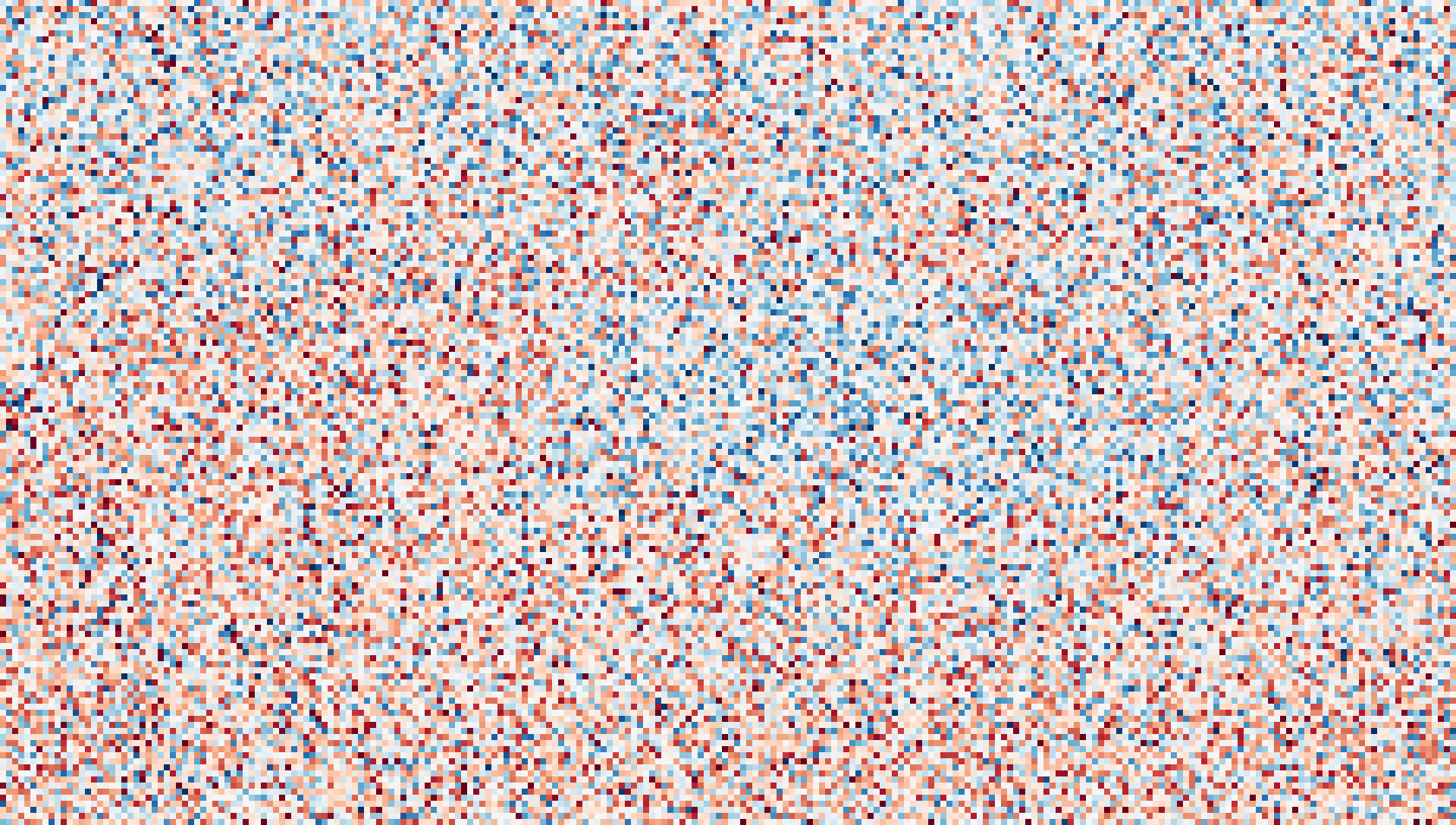}\hfill
    \includegraphics[width=.245\linewidth]{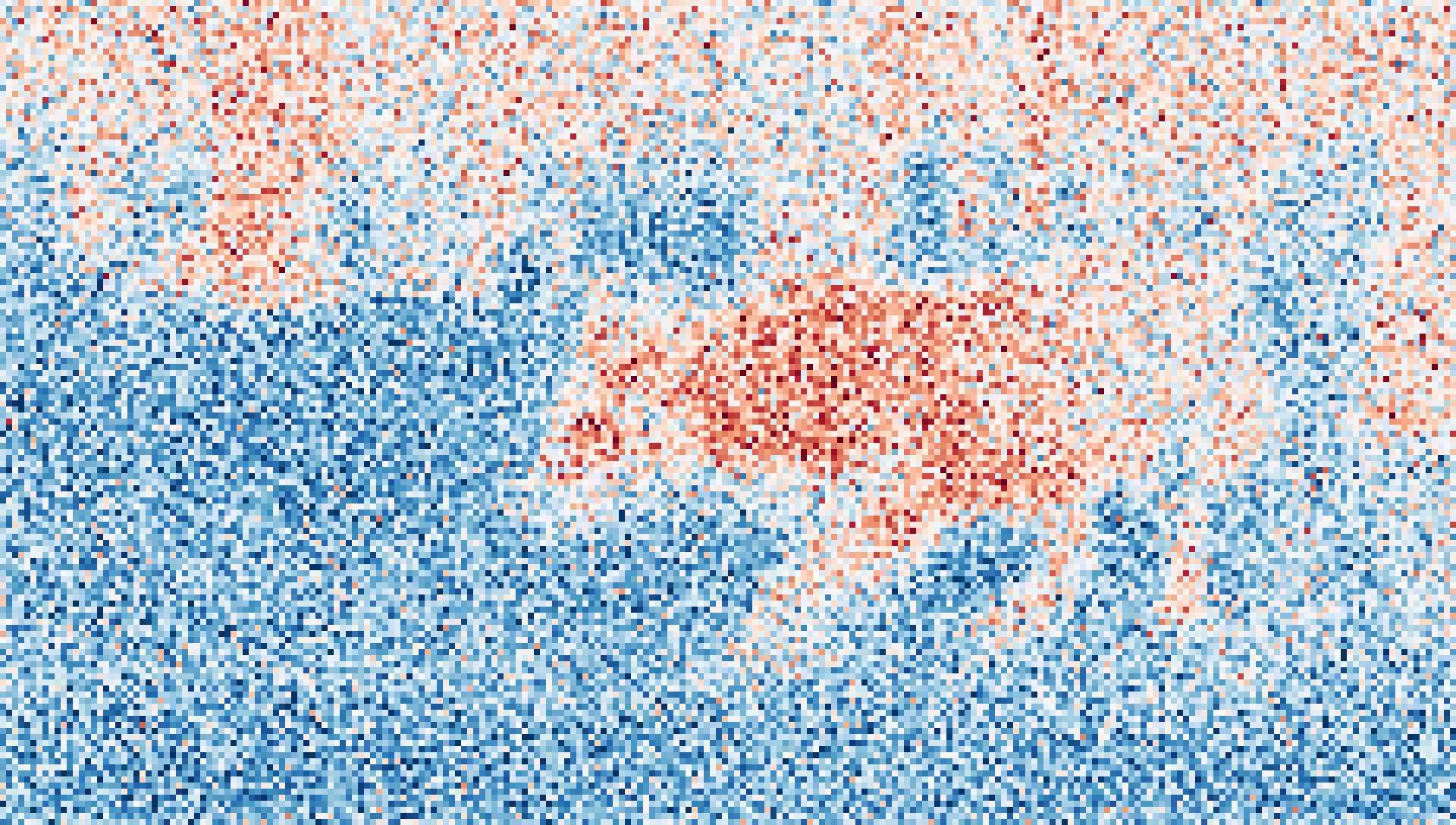}\hfill
    \includegraphics[width=.245\linewidth]{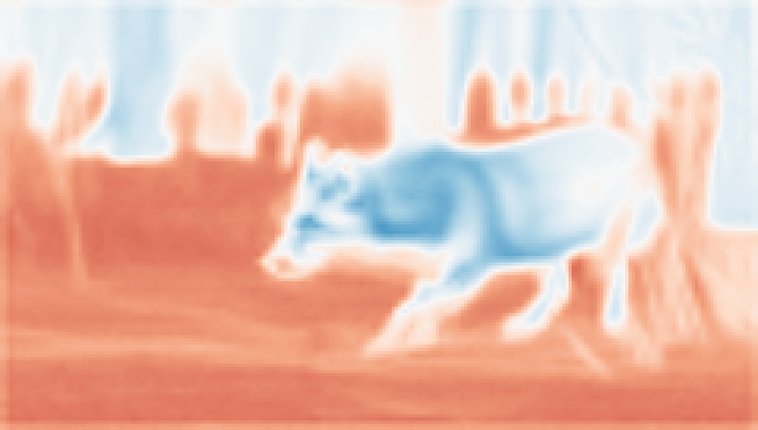}\hfill
    \includegraphics[width=.245\linewidth]{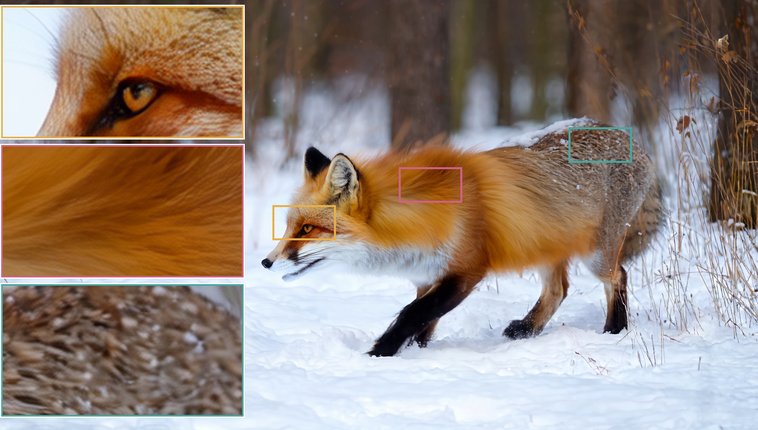}}%
  \par\nointerlineskip\vspace{1.5pt}
  \noindent\hbox to\linewidth{%
    \makebox[.245\linewidth]{\scriptsize\strut (e) High state}\hfill
    \makebox[.245\linewidth]{\scriptsize\strut (f) Corrected velocity}\hfill
    \makebox[.245\linewidth]{\scriptsize\strut (g) Aligned Clean prediction}\hfill
    \makebox[.245\linewidth]{\scriptsize\strut (h) Aligned result}}%
  \par
  \caption{Effect of trajectory synchronization during early denoising.
    Directly upsampling the low-resolution velocity (top) misaligns the
    clean prediction. The bottom row constructs the velocity from Low's clean
    prediction and High's current noisy state. This aligns their clean predictions
    while retaining High's own noisy state. Panels (a--c, e--g) visualize latents;
    (d, h) show the corresponding decoded results.}
  \label{fig:warmup-alignment}
\end{figure*}

\subsection{Overview}
\label{sec:overview}
TRACE extends pretrained video generation to higher resolutions without
training (Fig.~\ref{fig:method-overview}). Low-resolution generation
(Low, superscript $L$) first completes $N$ sampling steps, caching clean
predictions to guide high-resolution generation (High, $H$).
To share coarse structure from the start, we couple their initial latents
$z_0^L,z_0^H$ using independent standard Gaussian samples
$\epsilon^L,\epsilon^H$. During early High sampling, CSTS uses Low's
predictions to construct the velocity $v_p$ that updates the noisy latent
$z_p$ at step $p$ and noise level $\sigma_p$, avoiding High model calls.
High then refines local detail on a shared latent canvas, while TGSC
uses Low's layout to correct repeated structures. For larger resolutions,
we transfer High's refined result and continue generation using Low's
references, preserving scene structure and previously generated detail.

\pagebreak

\begingroup
\setlength{\abovedisplayskip}{5pt plus 1pt minus 1pt}
  \setlength{\belowdisplayskip}{5pt plus 1pt minus 1pt}
  \setlength{\abovedisplayshortskip}{5pt plus 1pt minus 1pt}
  \setlength{\belowdisplayshortskip}{5pt plus 1pt minus 1pt}
  \setlength{\jot}{2pt}
\subsection{Cross-Scale Trajectory Synchronization}
\label{sec:continuous-scale}

CSTS couples the initial noise across resolutions before constructing
High's early trajectory from Low's saved predictions. Let $R$ be a
whitened spatial downsampling
operator from High to Low, satisfying $RR^\top=I_L$.
Here $\top$ denotes transpose and $I_L,I_H$ are identity operators.
Using the independent samples $\epsilon^L,\epsilon^H$, Low starts at $z_0^L=\epsilon^L$ and High at the following coupled state
\begin{equation}
  z_0^H=R^\top\epsilon^L+(I_H-R^\top R)\epsilon^H.
  \label{eq:coupled-noise}
\end{equation}
The second term lies in the null space of $R$ and is independent of $\epsilon^L$.
Thus $Rz_0^H=\epsilon^L$ shares coarse noise without reducing
High's variance. High remains standard Gaussian, with covariance
$\operatorname{Cov}(z_0^H)=I_H$. Appendix~\textcolor{cvprblue}{A}
constructs the restriction operator and proves these noise properties.

\begin{figure*}[t]
  \centering
  \begin{minipage}[t]{.43\linewidth}\centering
    \includegraphics[width=\linewidth]{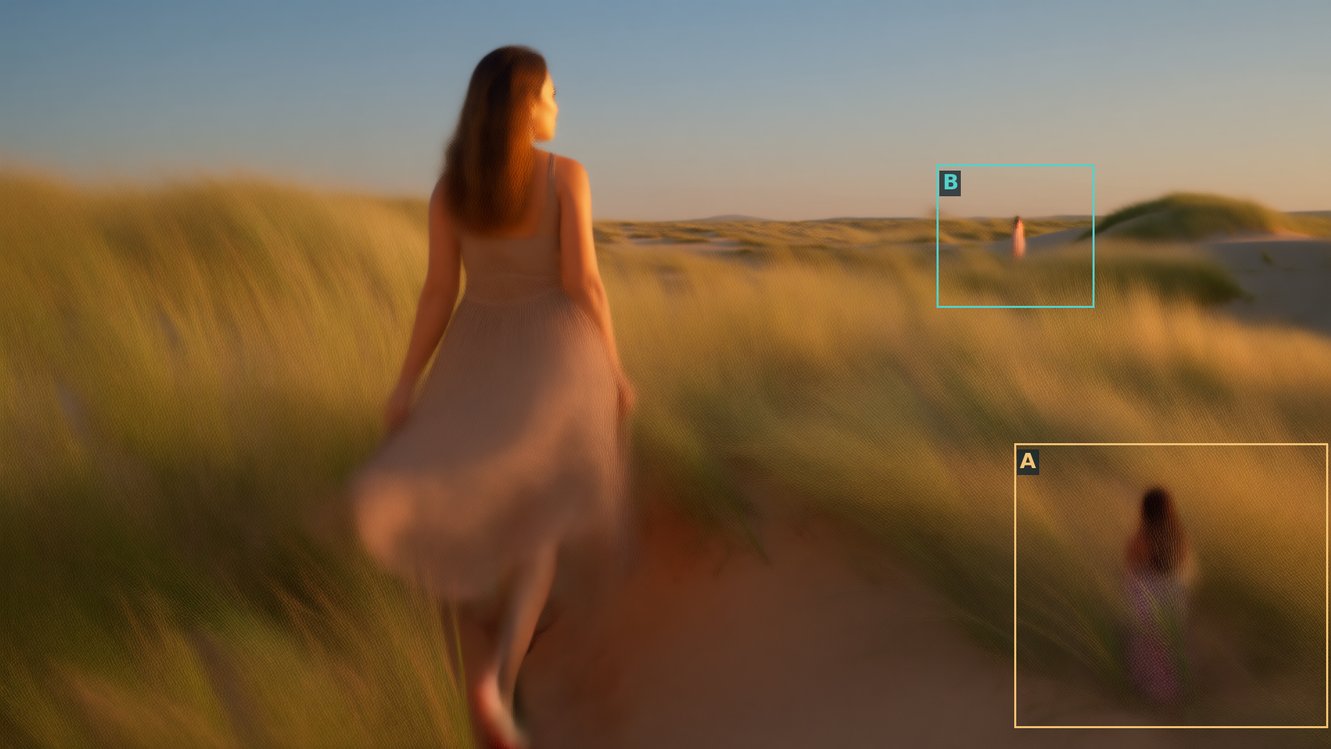}\par\vspace{2pt}
    {\fontsize{7pt}{8pt}\selectfont\textbf{(a) High-resolution prediction}}
  \end{minipage}\hfill
  \begin{minipage}[t]{.55\linewidth}\centering
    \begin{minipage}[t]{.242\linewidth}\centering
      \includegraphics[width=\linewidth]{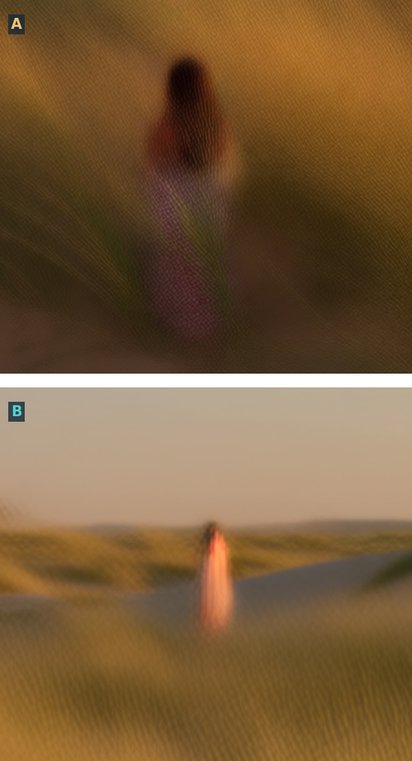}\par\vspace{2pt}
      {\fontsize{7pt}{8pt}\selectfont\textbf{(b) Before correction}}
    \end{minipage}\hfill
    \begin{minipage}[t]{.242\linewidth}\centering
      \includegraphics[width=\linewidth]{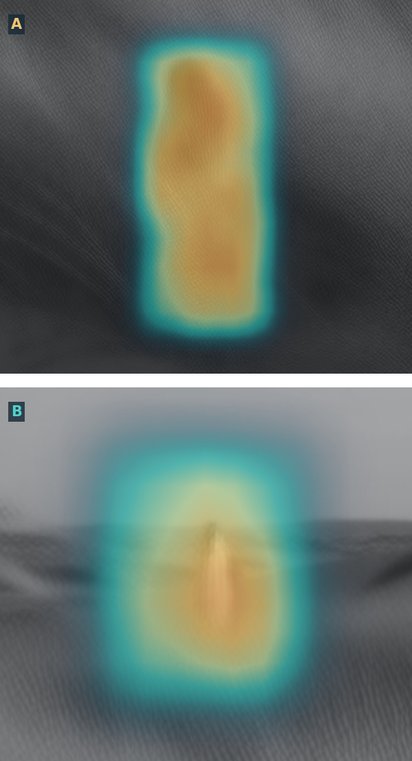}\par\vspace{2pt}
      {\fontsize{7pt}{8pt}\selectfont\textbf{(c) Detection evidence}}
    \end{minipage}\hfill
    \begin{minipage}[t]{.242\linewidth}\centering
      \includegraphics[width=\linewidth]{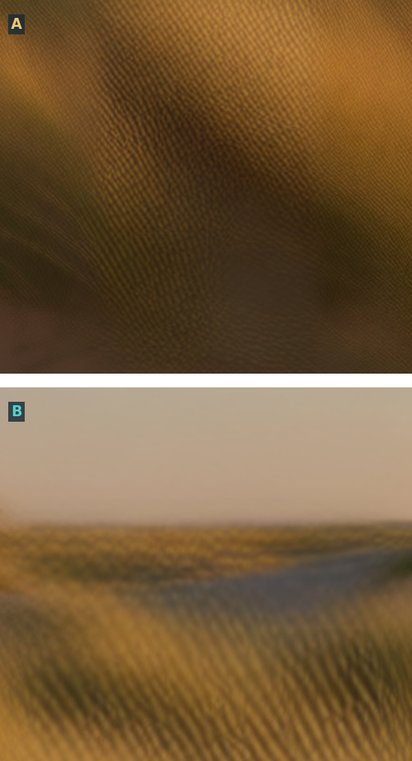}\par\vspace{2pt}
      {\fontsize{7pt}{8pt}\selectfont\textbf{(d) After correction}}
    \end{minipage}\hfill
    \begin{minipage}[t]{.242\linewidth}\centering
      \includegraphics[width=\linewidth]{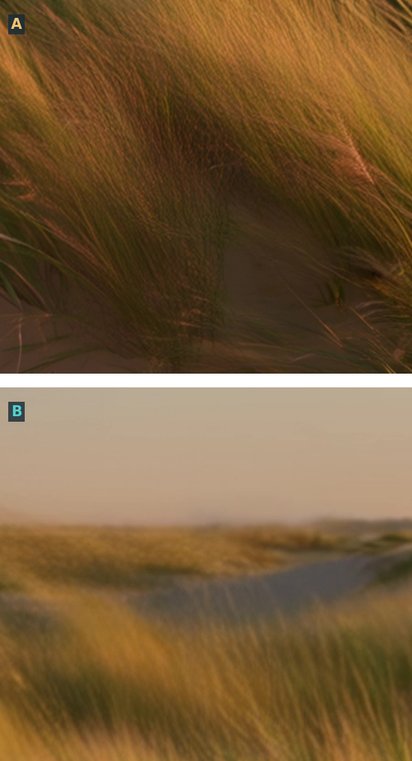}\par\vspace{2pt}
      {\fontsize{7pt}{8pt}\selectfont\textbf{(e) Final}}
    \end{minipage}
  \end{minipage}
  \vspace{-2pt}
  \caption{Trajectory-guided structural correction during high-resolution refinement.
    The prediction (a) contains two extra people absent from the low-resolution
    reference. Matched crops show these duplicated subjects (b), the evidence
    used to locate them (c), and their removal through regional correction (d).
    Later denoising adds grass detail in the corrected regions (e).}
  \vspace{-8.53581pt}
  \label{fig:regional-repair-illustration}
\end{figure*}

Shared initial noise alone does not align later predictions.
Increasing $p$ decreases noise: $1=\sigma_0>\cdots>\sigma_N=0$.
Low's velocity gives the clean prediction $\tilde z_p^L=z_p^L-\sigma_pv_p^L$.
Let $\mathcal U_H$ bilinearly upsample Low to High. Using Low's upsampled
velocity in High produces a clean prediction with a noisy-state mismatch:
\begin{equation}
  z_p^H-\sigma_p\mathcal U_Hv_p^L=\mathcal U_H\tilde z_p^L+\underbrace{(z_p^H-\mathcal U_Hz_p^L)}_{\text{noisy-state mismatch}}.
  \label{eq:main-unaligned-clean}
\end{equation}
The extra term shifts High's clean prediction away from Low's content.
We therefore use Low's clean prediction as the reference and construct
a velocity for High's current noisy state at each early denoising step.

The reference must also account for how content is transferred between
latent grids. VAE transfer $\mathcal T_H$ decodes a latent, upsamples the
frames, and re-encodes them. Its difference from latent upsampling $\mathcal U_H$
provides a fixed offset for the clean reference on the high-resolution grid.

\pagebreak
Let $s_H$ be High's first model step and $\tilde z_{s_H-1}^L$ Low's cached
clean prediction at the preceding step. We compute the fixed transfer offset
$\delta^H$ once before High starts and define its step-$p$ clean reference
$\bar z_p^H$ by
\begin{equation}
  \delta^H=(\mathcal T_H-\mathcal U_H)(\tilde z_{s_H-1}^L),\quad
  \bar z_p^H=\mathcal U_H\tilde z_p^L+\delta^H.
  \label{eq:aligned-reference}
\end{equation}

At step $p$, $z_p^H$ is the input state, obtained from coupled noise
at $p=0$ or the preceding update. For $p<s_H$, High's sampling
velocity $v_p^H$ uses Low's upsampled velocity with corrections for
the state mismatch and fixed transfer offset:
\begin{equation}
  v_p^H=\mathcal U_Hv_p^L+
  \frac{z_p^H-\mathcal U_Hz_p^L-\delta^H}{\sigma_p},\quad p<s_H.
  \label{eq:warm-flow}
\end{equation}
The calibration compensates for the difference between direct latent
upsampling and VAE transfer (Fig.~\ref{fig:warmup-alignment}). At the
pre-switch reference step, adding the measured offset reproduces the
VAE-transferred clean latent. Reusing this offset at earlier steps carries
the transfer adjustment through the trajectory without repeated VAE
evaluations. Together with the noisy-state correction, this lets High
inherit Low's evolving scene structure in a calibrated latent representation,
while advancing its own state without high-resolution model calls.

For $p=s_H,\ldots,N-1$, High obtains $v_p^H$ from its model using
standard classifier-free guidance (CFG).
Given $z_p^H$ and $v_p^H$, UniPC $\mathcal S_p$ updates the state
and incoming history $\mathcal H_p$ to the next step $p+1$:
\begin{equation}
  (z_{p+1}^H,\mathcal H_{p+1})
  =\mathcal S_p(z_p^H,v_p^H;\mathcal H_p).
  \label{eq:solver-transition}
\end{equation}
Switching velocity
sources retains High's state and history, without fresh noise or a solver
restart. High then adds detail to the scene built by early synchronization.
Its stored clean predictions remain available to later solver updates.
Appendix~\textcolor{cvprblue}{B} describes reference matching and
UniPC solver history at the switch.

\pagebreak

\subsection{Shared-Canvas Local Generation}
\label{sec:shared-canvas}

After synchronization, High uses local attention to refine the scene.
All windows share one latent state and one solver history across all video frames.

We apply local attention~\citep{liu2021swin,zhang2025sta,wu2026freeswim} on a patch grid
with $N_\tau$ frames, height $N_y$, and width $N_x$. A query at frame $\tau$ and
spatial position $(i_y,i_x)$ attends to a window $\mathcal W(i_y,i_x)$ of height
$w_y$ and width $w_x$ across all frame indices $0,\ldots,N_\tau-1$.
Its attention support $\mathcal V(\tau,i_y,i_x)$ is
\begin{equation}
  \mathcal V(\tau,i_y,i_x)=\{0,\ldots,N_\tau-1\}\times\mathcal W(i_y,i_x).
  \label{eq:window-support}
\end{equation}
Windows shift inward at borders while preserving global coordinates.
For query, key, and value vectors $\mathbf q_i,\mathbf k_j,\mathbf v_j$
and head width $d$, the attention output is
\begin{equation}
  \mathbf h_i'=\sum_{j\in\mathcal V(i)}
  \operatorname{softmax}_{j\in\mathcal V(i)}\!\left(
  \frac{\mathbf q_i^\top\mathbf k_j}{\sqrt d}\right)\mathbf v_j.
  \label{eq:local-attention}
\end{equation}

Here $i,j$ index tokens, softmax is normalized over the indicated support, and $\mathcal V(i)$ is the support in
Eq.~\eqref{eq:window-support} for query $i$. Each query attends to $N_\tau w_y w_x$ keys, reducing
attention cost from $O(N_{\mathrm{tok}}^2d)$ to $O(N_{\mathrm{tok}}N_\tau w_y w_x d)$
for $N_{\mathrm{tok}}=N_\tau N_y N_x$ tokens.
The windows retain temporal context across all latent frames, while
overlapping supports exchange spatial information across layers.
Appendix~\textcolor{cvprblue}{C} details window placement near spatial boundaries.

Local windows see only part of the scene and can create extra subjects.
We use Low's layout to detect and correct these errors in scene structure in the following section.

\subsection{Trajectory-Guided Structural Correction}
\label{sec:consensus-tube}

\begin{table*}[t]
  \vspace{-10.243pt}
  \caption{Quantitative results of different methods for high-resolution video generation at
    $1920\!\times\!1088$ on Wan2.1-T2V-1.3B and Bernini-R-1.3B, and
    $2560\!\times\!1440$ on Wan2.2-T2V-A14B.
    Scores measure video quality, temporal consistency, text alignment,
    patch-level image statistics, and total inference time.
    \textcolor{TableBest}{Red} and \textcolor{TableSecond}{blue} mark the
    best and second-best method scores.}
  \vspace{-8.53581pt}
  \label{tab:main_full}
  \centering
  \scriptsize
  \setlength{\tabcolsep}{0.85pt}
  \renewcommand{\arraystretch}{0.90}
  \begin{tabular*}{\linewidth}{@{\extracolsep{\fill}}l*{14}{c}@{}}
    \toprule
    \multirow{2}{*}{\textbf{Method}}
    & \multicolumn{3}{c}{\textbf{Video Quality}}
    & \multicolumn{7}{c}{\textbf{VBench}}
    & \multicolumn{1}{c}{\textbf{Alignment}}
    & \multicolumn{2}{c}{\textbf{HR Fidelity}}
    & \multicolumn{1}{c}{\textbf{Latency}} \\
    \cmidrule(lr){2-4}\cmidrule(lr){5-11}\cmidrule(lr){12-12}
    \cmidrule(lr){13-14}\cmidrule(l){15-15}
    & QA$\uparrow$ & DOVER$\uparrow$ & VS$\uparrow$
    & SC$\uparrow$ & MS$\uparrow$ & BC$\uparrow$ & OC$\uparrow$
    & AQ$\uparrow$ & IQ$\uparrow$ & Mean$\uparrow$
    & VQA$\uparrow$ & BRISQUE$^{*}\downarrow$ & NIQE$^{*}\downarrow$
    & Time (s)$\downarrow$ \\
    \midrule
    \rowcolor{TableGroup}
    \multicolumn{15}{c}{\textbf{Wan2.1-T2V-1.3B~\citep{wanteam2025wan}\quad $1920\!\times\!1088$}} \\
    {\fontsize{6.3pt}{6.8pt}\selectfont Direct Inference}
    & 4.165 & 0.5875 & 2.730
    & \textcolor{TableBest}{\textbf{0.9787}} & 0.9855 & \textcolor{TableSecond}{\textbf{0.9755}} & 0.2224 & 0.6008 & 0.6570 & 0.7366
    & 0.6878 & 0.3400 & 0.2060 & 3,366 \\
    {\fontsize{6.3pt}{6.8pt}\selectfont C.Scaling~\citep{guo2024cheapscaling}}
    & \textcolor{TableSecond}{\textbf{4.568}} & \textcolor{TableSecond}{\textbf{0.6546}} & 2.747
    & 0.9690 & 0.9810 & 0.9732 & 0.2400 & \textcolor{TableSecond}{\textbf{0.6454}} & 0.7111 & 0.7533
    & 0.8539 & 0.3518 & 0.2063 & 3,927 \\
    {\fontsize{6.3pt}{6.8pt}\selectfont FreeScale~\citep{qiu2025freescale}}
    & 3.761 & 0.5835 & 2.718
    & 0.9676 & 0.9798 & 0.9707 & 0.2344 & 0.6010 & 0.6435 & 0.7328
    & 0.8207 & 0.4282 & 0.2579 & 3,968 \\
    {\fontsize{6.3pt}{6.8pt}\selectfont RIFLEx~\citep{zhao2025riflex}}
    & 4.080 & 0.5840 & 2.800
    & \textcolor{TableSecond}{\textbf{0.9753}} & 0.9857 & 0.9735 & 0.2192 & 0.5874 & 0.6420 & 0.7305
    & 0.6620 & 0.3380 & 0.2570 & 3,338 \\
    {\fontsize{6.3pt}{6.8pt}\selectfont CineScale~\citep{qiu2025cinescale}}
    & 3.854 & 0.6234 & 2.725
    & 0.9681 & 0.9796 & 0.9736 & 0.2388 & 0.6085 & 0.6686 & 0.7395
    & 0.8630 & 0.3406 & 0.2071 & \textcolor{TableSecond}{\textbf{2,360}} \\
    {\fontsize{6.3pt}{6.8pt}\selectfont FreeSwim~\citep{wu2026freeswim}}
    & 4.381 & 0.5525 & \textcolor{TableSecond}{\textbf{2.833}}
    & 0.9705 & \textcolor{TableBest}{\textbf{0.9901}} & 0.9753 & \textcolor{TableBest}{\textbf{0.2514}} & 0.6377 & \textcolor{TableSecond}{\textbf{0.7172}} & \textcolor{TableSecond}{\textbf{0.7570}}
    & \textcolor{TableSecond}{\textbf{0.8735}} & \textcolor{TableSecond}{\textbf{0.2132}} & \textcolor{TableSecond}{\textbf{0.1534}} & 5,010 \\
    {\fontsize{6.3pt}{6.8pt}\selectfont \textcolor{QualAccent}{\textbf{TRACE (Ours)}}}
    & \textcolor{TableBest}{\textbf{4.690}} & \textcolor{TableBest}{\textbf{0.6857}} & \textcolor{TableBest}{\textbf{2.856}}
    & 0.9707 & \textcolor{TableSecond}{\textbf{0.9899}} & \textcolor{TableBest}{\textbf{0.9791}} & \textcolor{TableSecond}{\textbf{0.2492}} & \textcolor{TableBest}{\textbf{0.6648}} & \textcolor{TableBest}{\textbf{0.7193}} & \textcolor{TableBest}{\textbf{0.7622}}
    & \textcolor{TableBest}{\textbf{0.8870}} & \textcolor{TableBest}{\textbf{0.1897}} & \textcolor{TableBest}{\textbf{0.1406}} & \textcolor{TableBest}{\textbf{1,826}} \\
    \midrule
    \rowcolor{TableGroup}
    \multicolumn{15}{c}{\textbf{Wan2.2-T2V-A14B~\citep{wanteam2025wan}\quad $2560\!\times\!1440$}} \\

    {\fontsize{6.3pt}{6.8pt}\selectfont FreeSwim~\citep{wu2026freeswim}}
    & 4.524 & 0.6084 & 2.757
    & 0.9397 & \textcolor{TableBest}{\textbf{0.9848}} & 0.9513 & \textcolor{TableBest}{\textbf{0.2562}} & 0.6647 & \textcolor{TableBest}{\textbf{0.7337}} & 0.7551
    & 0.9160 & \textcolor{TableBest}{\textbf{0.1736}} & 0.1300 & $ 27{,}788$ \\
    {\fontsize{6.3pt}{6.8pt}\selectfont \textcolor{QualAccent}{\textbf{TRACE (Ours)}}}
    & \textcolor{TableBest}{\textbf{4.752}} & \textcolor{TableBest}{\textbf{0.6900}} & \textcolor{TableBest}{\textbf{2.793}}
    & \textcolor{TableBest}{\textbf{0.9673}} & 0.9846 & \textcolor{TableBest}{\textbf{0.9722}} & 0.2544 & \textcolor{TableBest}{\textbf{0.6656}} & 0.7324 & \textcolor{TableBest}{\textbf{0.7628}}
    & \textcolor{TableBest}{\textbf{0.9519}} & 0.1817 & \textcolor{TableBest}{\textbf{0.1294}} & \textcolor{TableBest}{\textbf{9,526}} \\
    \midrule
    \rowcolor{TableGroup}
    \multicolumn{15}{c}{\textbf{Bernini-R-1.3B~\citep{liu2026bernini}\quad $1920\!\times\!1088$}} \\
    {\fontsize{6.3pt}{6.8pt}\selectfont FreeSwim~\citep{wu2026freeswim}}
    & 4.572 & 0.6017 & 2.734
    & 0.9698 & 0.9892 & 0.9727 & \textcolor{TableBest}{\textbf{0.2548}} & 0.6553 & \textcolor{TableBest}{\textbf{0.7216}} & 0.7606
    & 0.8929 & 0.1941 & 0.1349 & 5,708 \\
    {\fontsize{6.3pt}{6.8pt}\selectfont \textcolor{QualAccent}{\textbf{TRACE (Ours)}}}
    & \textcolor{TableBest}{\textbf{4.727}} & \textcolor{TableBest}{\textbf{0.6946}} & \textcolor{TableBest}{\textbf{2.812}}
    & \textcolor{TableBest}{\textbf{0.9740}} & \textcolor{TableBest}{\textbf{0.9895}} & \textcolor{TableBest}{\textbf{0.9780}} & 0.2517 & \textcolor{TableBest}{\textbf{0.6723}} & 0.7171 & \textcolor{TableBest}{\textbf{0.7637}}
    & \textcolor{TableBest}{\textbf{0.9024}} & \textcolor{TableBest}{\textbf{0.1860}} & \textcolor{TableBest}{\textbf{0.1316}} & \textcolor{TableBest}{\textbf{2,831}} \\
    \bottomrule
  \end{tabular*}
  \vspace{-1.42262pt}
\end{table*}

Local refinement can create both useful detail and extra subjects.
TGSC uses Low's layout to identify extra subjects and correct the affected regions
(Fig.~\ref{fig:regional-repair-illustration}).

We first use cross-attention relevance $C_p^n$ to locate regions associated
with prompt entity $n$ at denoising step $p$. These regions may contain
either the intended subject or an extra instance elsewhere in the scene.

Self-attention scores $D_p$ and $V_p$ measure High--Low disagreement
and High's change from step $p-1$, respectively.
Here $\Phi(\cdot)$ maps each signal's ascending spatial ranks within
each frame to $[0,1]$. With $\odot$ denoting elementwise multiplication,
we combine the normalized signals:
\begin{equation}
  S^n_p=\left[\Phi(C^n_p)\odot
    \Phi(D_p)\odot\Phi(V_p)\right]^{1/3}.
  \label{eq:deviation-score}
\end{equation}
The combined score favors locations that respond to the entity, differ
from Low's structure, and change during High's refinement. We retain
high-scoring locations from $p-1$ where the current prediction still
responds to the same entity, allowing small spatial shifts. Grouping
these locations gives candidate regions for further checks.

To verify each candidate, we use Low's cross-attention to identify
locations that respond strongly to the target entity and distant locations
with weak responses. These responses provide a reference for judging
whether a candidate in High responds like the target entity at the same
noise level. Each candidate must show strong entity
evidence and agreement across attention heads at both $p-1$ and $p$,
filtering out transient attention changes during refinement.

We next determine whether the detected region should be preserved.
Low's original subject region is protected, and candidates better
explained by another prompt entity are excluded. The remaining verified
regions form $\mathcal R_p$. Appendix~\textcolor{cvprblue}{D} specifies
the attention scores, candidate construction, and semantic checks.

The selected regions must also remain connected across video frames.
The mask operator $\mathcal F$ links their boxes across sampled frames,
interpolates their positions, and softens their boundaries. Overlapping
boxes are merged into one mask so each location receives a single correction.
TGSC is invoked at every High model step. The correction mask is
\begin{equation}
  M_p=\mathcal F(\mathcal R_p),\qquad \mathcal F(\varnothing)=0.
  \label{eq:mask-spatial}
\end{equation}
An empty region set leaves the latent unchanged.

\pagebreak
Replacing a region with clean content would remove noise still needed
by the sampler. Let $r_\epsilon=z_0^H-\mathcal U_Hz_0^L$ be the
initial latent difference, and let $z^L(\sigma_p)=z_p^L$ retrieve Low's state
by noise level. The fixed transfer calibration and initial noise
residual give the noise-matched reference
\begin{equation}
  T_\sigma=\mathcal U_Hz^L(\sigma)+(1-\sigma)\delta^H+\sigma r_\epsilon.
  \label{eq:repair-target}
\end{equation}
At $\sigma=1$, this reference recovers High's initial noise; at
$\sigma=0$, it becomes Low's calibrated endpoint. This preserves
the noise level while transferring Low's structure. We correct
the solver output at $\sigma_{p+1}$:
\begin{equation}
  z_{p+1}^H\leftarrow
  (1-M_p)\odot z_{p+1}^H+M_p\odot T_{\sigma_{p+1}}.
  \label{eq:regional-repair}
\end{equation}
The assignment updates only masked regions of the solver output,
with a smooth transition at the boundary.

UniPC's stored predictions and states must remain consistent with the
corrected content. We apply the same mask using matching references.
Appendix~\textcolor{cvprblue}{E} details the matched references
and consistent updates of stored states and clean predictions.

\subsection{Extension to Ultra-High Resolution}
\label{sec:second-scale}
For ultra-high resolution ($U$), we extend the procedure to a larger grid,
using High as an intermediate resolution. Latent upsampling $\mathcal U_U$ and VAE transfer
$\mathcal T_U$ now map High to $U$. The key is to retain High's completed
result while reusing Low's cached denoising trajectory:
\begin{equation}
  \begin{aligned}
    \delta^U & =\mathcal T_U(z_N^H)-\mathcal U_U\mathcal U_Hz_N^L, \\
    v_p^U & =\mathcal U_U\mathcal U_Hv_p^L+
    \frac{z_p^U-\mathcal U_U\mathcal U_Hz_p^L-\delta^U}{\sigma_p},\quad p<s_U.
  \end{aligned}
  \label{eq:main-second-bridge}
\end{equation}
Here $s_U$ is the first model step; sampling otherwise follows CSTS.
The offset carries both VAE calibration and High's refinements beyond Low.
Using High as an intermediate stage therefore preserves the detail and
TGSC corrections developed there, giving the final stage a refined
reference before further scaling. Appendix~\textcolor{cvprblue}{F} derives
this progressive extension.

\par\endgroup

\pagebreak
\section{Experiments}
\label{sec:experiments}
\subsection{Experimental Setup}
\begin{figure*}[!t]
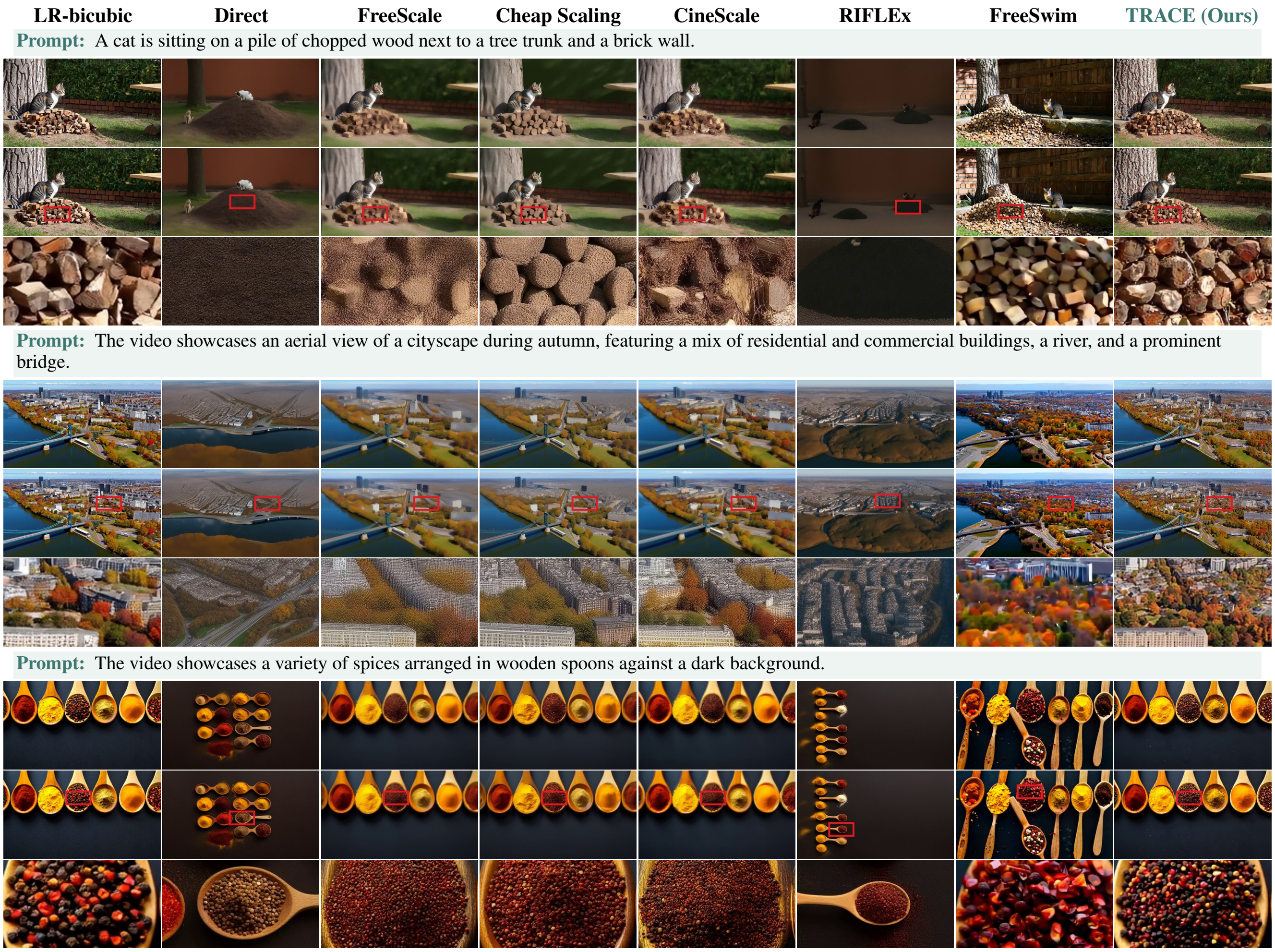

  \centering
  \setlength{\abovecaptionskip}{1pt}%
  \setlength{\belowcaptionskip}{0pt}%
  \setlength{\parskip}{0pt}
  \noindent\makebox[.125\linewidth]{{\fontsize{8pt}{9pt}\selectfont\textbf{LR-bicubic}}}%
  \makebox[.125\linewidth]{{\fontsize{8pt}{9pt}\selectfont\textbf{Direct}}}%
  \makebox[.125\linewidth]{{\fontsize{8pt}{9pt}\selectfont\textbf{FreeScale}}}%
  \makebox[.125\linewidth]{{\fontsize{8pt}{9pt}\selectfont\textbf{Cheap Scaling}}}%
  \makebox[.125\linewidth]{{\fontsize{8pt}{9pt}\selectfont\textbf{CineScale}}}%
  \makebox[.125\linewidth]{{\fontsize{8pt}{9pt}\selectfont\textbf{RIFLEx}}}%
  \makebox[.125\linewidth]{{\fontsize{8pt}{9pt}\selectfont\textbf{FreeSwim}}}%
  \makebox[.125\linewidth]{{\fontsize{8pt}{9pt}\selectfont\textbf{\textcolor{QualAccent}{TRACE (Ours)}}}}\par\vspace{1pt}\nointerlineskip
  \noindent%
  \begingroup
  \setlength{\fboxsep}{1.2pt}%
  \colorbox{PromptBg}{\parbox[c]{.98\linewidth}{%
      \raggedright\fontsize{7.5pt}{8.3pt}\selectfont
      \textcolor{QualAccent}{\textbf{Prompt:}}\enspace A cat is sitting on a pile of chopped wood next to a tree trunk and a brick wall.\strut}}%
  \endgroup\par\nointerlineskip\vspace{1pt}%
  \noindent\makebox[\linewidth][c]{%
    \input{figures/visuals/comparisons/latex/wan21_1080/HR-T2V-0005.tex}}\par\nointerlineskip
  \vspace{2pt}
  \noindent%
  \begingroup
  \setlength{\fboxsep}{1.2pt}%
  \colorbox{PromptBg}{\parbox[c]{.98\linewidth}{%
      \raggedright\fontsize{7.5pt}{8.3pt}\selectfont
      \textcolor{QualAccent}{\textbf{Prompt:}}\enspace The video showcases an aerial view of a cityscape during autumn, featuring a mix of residential and commercial buildings, a river, and a prominent bridge.\strut}}%
  \endgroup\par\nointerlineskip\vspace{1pt}%
  \noindent\makebox[\linewidth][c]{%
    \input{figures/visuals/comparisons/latex/wan21_1080/HR-T2V-0019.tex}}\par\nointerlineskip
  \vspace{2pt}
  \noindent%
  \begingroup
  \setlength{\fboxsep}{1.2pt}%
  \colorbox{PromptBg}{\parbox[c]{.98\linewidth}{%
      \raggedright\fontsize{7.5pt}{8.3pt}\selectfont
      \textcolor{QualAccent}{\textbf{Prompt:}}\enspace The video showcases a variety of spices arranged in wooden spoons against a dark background.\strut}}%
  \endgroup\par\nointerlineskip\vspace{1pt}%
  \noindent\makebox[\linewidth][c]{%
    \input{figures/visuals/comparisons/latex/wan21_1080/HR-T2V-0007.tex}}\par\nointerlineskip
  \caption{Visual comparison of direct inference and training-free
    resolution-extension methods at $1920\!\times\!1088$ on Wan2.1.
    Each result pairs two adjacent frames with a detail crop, highlighting
    wood texture, building facades, and local structure.}
  \label{fig:qual-wan21-1088}
  \vspace{-8.53581pt}
\end{figure*}
\noindent\textbf{Benchmark.}
Our test set uses 120 fixed prompt--reference pairs selected from
UltraVideo~\citep{xue2025ultravideo}, following the 4K-VBench protocol of
T3-Video~\citep{zhang2025t3video}. Wan2.1-T2V-1.3B~\citep{wanteam2025wan} and
Bernini-R-1.3B~\citep{liu2026bernini} generate $1920\!\times\!1088$ videos, with Low at
$832\!\times\!480$ and $848\!\times\!480$, respectively. Wan2.2-T2V-A14B~\citep{wanteam2025wan}
uses $1280\!\times\!720$ for Low and $2560\!\times\!1440$ for High. All methods
generate 81 frames at 16~FPS with matched seeds and the fixed Wan
negative prompt. Table~\ref{tab:main_full} reports end-to-end latency on one
A800 GPU.

\noindent\textbf{Comparison methods.}
On Wan2.1, we compare direct inference, Cheap
Scaling~\citep{guo2024cheapscaling}, FreeScale~\citep{qiu2025freescale},
RIFLEx~\citep{zhao2025riflex}, CineScale~\citep{qiu2025cinescale}, and
FreeSwim~\citep{wu2026freeswim}. On Wan2.2 and Bernini, we focus on
FreeSwim~\citep{wu2026freeswim}, a state-of-the-art training-free baseline that also uses
window attention. We do not include guided samplers developed for
text-to-image~\citep{bu2025hiflow}, image-to-video~\citep{casellesdupre2026fresco},
or tiled generation~\citep{koroglu2026jolt}, as adapting them to this
benchmark requires additional implementation choices.
Appendix~\textcolor{cvprblue}{G} compares TRACE with VSR methods
applied to the same low-resolution outputs.

\noindent\textbf{Implementation details.}
All backbone weights remain frozen. Wan2.1 uses 50 denoising steps with
flow shift 5; Wan2.2 and Bernini use 40 steps with shifts 12 and 3; Low and High
use noise levels from the same shared schedule. We use the same TRACE
procedure without per-video tuning. Appendix~\textcolor{cvprblue}{F}
describes the inference order, solver handoff, and backbone-specific
guidance settings.

\noindent\textbf{Evaluation.}
We assess video quality with Q-Align (QA)~\citep{wu2024qalign},
DOVER-overall~\citep{wu2023dover}, and VideoScore visual quality
(VS)~\citep{he2024videoscore}, and text alignment with VQAScore
(VQA)~\citep{lin2024vqascore}. VBench~\citep{huang2024vbench} reports
subject consistency (SC), motion smoothness (MS), background consistency
(BC), overall consistency (OC), aesthetic quality (AQ), and imaging
quality (IQ); Mean is their average. We compute normalized
BRISQUE~\citep{mittal2012brisque} and NIQE~\citep{mittal2013niqe}
score gaps between generated and reference patches:
\begingroup
\predisplaypenalty=0
\begin{equation}
  m^*=\frac{1}{K}\sum_{k=1}^{K}
  \min\!\left\{\frac{|m(P_k)-m(Q_k)|}{\rho_m},\,1\right\}.
  \label{eq:detail-scores}
\end{equation}
\endgroup
Here $m$ is BRISQUE or NIQE; $P_k,Q_k$ are generated/reference
patches at $K$ matched normalized time--space
positions. The normalization $\rho_m$ is the 1st--99th percentile
range of reference scores at each resolution. Smaller gaps indicate closer local image statistics rather than a
pixelwise match between generated and reference content.

\flushbottom

\begin{figure}[!t]
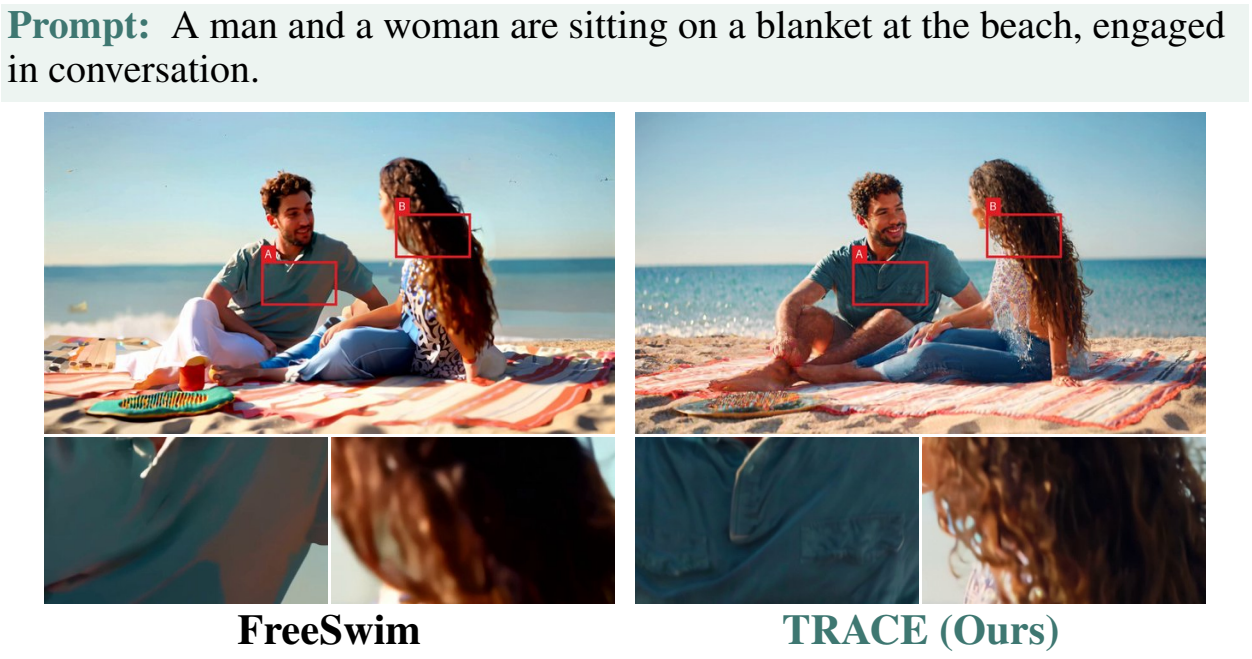

  \centering
  \setlength{\abovecaptionskip}{1pt}%
  \setlength{\belowcaptionskip}{0pt}%
  \setlength{\parskip}{0pt}
  \begingroup
  \setlength{\fboxsep}{1.2pt}%
  \colorbox{PromptBg}{\parbox[c]{.98\linewidth}{%
      \raggedright\fontsize{7.5pt}{8.3pt}\selectfont
      \textcolor{QualAccent}{\textbf{Prompt:}}\enspace A man and a woman are sitting on a blanket at the beach, engaged in conversation.\strut}}%
  \endgroup\par\nointerlineskip\vspace{2pt}%
  \noindent\hbox to\linewidth{\hfil
    \begin{minipage}[t]{107.32265pt}\centering
      \input{figures/visuals/comparisons/latex/wan21_2160/HR-T2V-0009/freeswim.tex}\par\nointerlineskip\vspace{2pt}%
      \hbox to\linewidth{\hfil{\fontsize{8pt}{9pt}\selectfont\textbf{FreeSwim}}\hfil}%
    \end{minipage}\hspace{4pt}%
    \begin{minipage}[t]{107.32265pt}\centering
      \input{figures/visuals/comparisons/latex/wan21_2160/HR-T2V-0009/ours.tex}\par\nointerlineskip\vspace{2pt}%
      \hbox to\linewidth{\hfil{\fontsize{8pt}{9pt}\selectfont\textbf{\textcolor{QualAccent}{TRACE (Ours)}}}\hfil}%
    \end{minipage}\hfil}\par
  \vspace{2pt}
  \captionof{figure}{Visual comparison of FreeSwim and TRACE at $3840{\times}2160$ on Wan2.1, showing full frames and enlarged crops.}%
\label{fig:qual-wan21-4k}
  \vspace{3pt}
\end{figure}
\subsection{Main Results}
\label{sec:results}

\noindent\textbf{Quantitative results.}
TRACE delivers better perceptual quality and text alignment across all
three backbones, leading in QA, DOVER, VS, and VQAScore
(Table~\ref{tab:main_full}). It also achieves the highest VBench Mean,
with more consistent subjects and backgrounds than FreeSwim. Its higher
aesthetic scores further support more pleasing visual appearance.
Generally smaller BRISQUE$^*$ and NIQE$^*$ gaps show closer agreement
with reference image statistics, supporting better local fidelity at the higher output resolutions.

\pagebreak
\noindent\textbf{Inference efficiency.}
TRACE is $2.74\times$, $2.92\times$, and $2.02\times$ faster than
FreeSwim on Wan2.1, Wan2.2, and Bernini. These end-to-end timings
include Low sampling, VAE transfers, and TGSC. The advantage grows
at ultra-high resolution: Fig.~\ref{fig:intro-visual} shows an $11.8\times$
speedup at $3840\!\times\!2160$ on Wan2.1. Skipping early high-resolution
inference saves more computation on larger grids.

\begin{figure*}[p]
  \setlength{\abovecaptionskip}{1pt}%
  \setlength{\belowcaptionskip}{0pt}%
  \setlength{\parskip}{0pt}
  \noindent\begin{minipage}[t][\dimexpr\textheight-\dp\strutbox\relax][s]{.48\textwidth}
    \vspace{0pt}
    \centering
    \begingroup
    \setlength{\fboxsep}{1.2pt}%
    \colorbox{PromptBg}{\parbox[c]{.98\linewidth}{%
        \raggedright\fontsize{7.5pt}{8.3pt}\selectfont
        \textcolor{QualAccent}{\textbf{Prompt:}}\enspace The video showcases a breathtaking aerial view of a majestic mountain surrounded by lush greenery and vast ocean waters.\strut}}%
    \endgroup\par\nointerlineskip\vspace{2pt}%
    \noindent\hbox to\linewidth{\hfil
      \begin{minipage}[t]{107.32265pt}\centering
        \input{figures/visuals/comparisons/latex/wan22_1440/HR-T2V-0018/freeswim.tex}\par\nointerlineskip\vspace{2pt}%
        \hbox to\linewidth{\hfil{\fontsize{8pt}{9pt}\selectfont\textbf{FreeSwim}}\hfil}%
      \end{minipage}\hspace{4pt}%
      \begin{minipage}[t]{107.32265pt}\centering
        \input{figures/visuals/comparisons/latex/wan22_1440/HR-T2V-0018/ours.tex}\par\nointerlineskip\vspace{2pt}%
        \hbox to\linewidth{\hfil{\fontsize{8pt}{9pt}\selectfont\textbf{\textcolor{QualAccent}{TRACE (Ours)}}}\hfil}%
      \end{minipage}\hfil}\par
    \par\nointerlineskip\vspace{4pt}
    \begingroup
    \setlength{\fboxsep}{1.2pt}%
    \colorbox{PromptBg}{\parbox[c]{.98\linewidth}{%
        \raggedright\fontsize{7.5pt}{8.3pt}\selectfont
        \textcolor{QualAccent}{\textbf{Prompt:}}\enspace The video shows a small outdoor fire burning within a brick enclosure.\strut}}%
    \endgroup\par\nointerlineskip\vspace{2pt}%
    \noindent\hbox to\linewidth{\hfil
      \begin{minipage}[t]{107.32265pt}\centering
        \input{figures/visuals/comparisons/latex/wan22_1440/HR-T2V-0097/freeswim.tex}\par\nointerlineskip\vspace{2pt}%
        \hbox to\linewidth{\hfil{\fontsize{8pt}{9pt}\selectfont\textbf{FreeSwim}}\hfil}%
      \end{minipage}\hspace{4pt}%
      \begin{minipage}[t]{107.32265pt}\centering
        \input{figures/visuals/comparisons/latex/wan22_1440/HR-T2V-0097/ours.tex}\par\nointerlineskip\vspace{2pt}%
        \hbox to\linewidth{\hfil{\fontsize{8pt}{9pt}\selectfont\textbf{\textcolor{QualAccent}{TRACE (Ours)}}}\hfil}%
      \end{minipage}\hfil}\par
    \captionof{figure}{FreeSwim and TRACE at $2560{\times}1440$ on Wan2.2.}%
\label{fig:qual-wan22-1440}
    \par\nointerlineskip\vspace{-1pt}
    \begingroup
    \setlength{\fboxsep}{1.2pt}%
    \colorbox{PromptBg}{\parbox[c]{.98\linewidth}{%
        \raggedright\fontsize{7.5pt}{8.3pt}\selectfont
        \textcolor{QualAccent}{\textbf{Prompt:}}\enspace A woman in a flowing dress walks barefoot across sandy dunes with tall grasses swaying in the background, bathed in the golden light of the setting sun.\strut}}%
    \endgroup\par\nointerlineskip\vspace{2pt}%
    \noindent\hbox to\linewidth{\hfil
      \begin{minipage}[t]{107.32265pt}\centering
        \input{figures/visuals/comparisons/latex/wan22_2160/HR-T2V-0010/freeswim.tex}\par\nointerlineskip\vspace{2pt}%
        \hbox to\linewidth{\hfil{\fontsize{8pt}{9pt}\selectfont\textbf{FreeSwim}}\hfil}%
      \end{minipage}\hspace{4pt}%
      \begin{minipage}[t]{107.32265pt}\centering
        \input{figures/visuals/comparisons/latex/wan22_2160/HR-T2V-0010/ours.tex}\par\nointerlineskip\vspace{2pt}%
        \hbox to\linewidth{\hfil{\fontsize{8pt}{9pt}\selectfont\textbf{\textcolor{QualAccent}{TRACE (Ours)}}}\hfil}%
      \end{minipage}\hfil}\par
    \captionof{figure}{FreeSwim and TRACE at $3840{\times}2160$ on Wan2.2. }%
\label{fig:qual-wan22-2160}
    \par\vspace{1pt}
    \leftskip=0pt\rightskip=0pt
    \normalfont\normalsize
    \flushbottom
  \setlength{\parskip}{0pt plus 1pt}
  \interlinepenalty=10000
  \predisplaypenalty=10000
  \postdisplaypenalty=10000
  \setlength{\parfillskip}{0pt plus .25\linewidth}

    \noindent\textbf{Visual comparison.}
    In Fig.~\ref{fig:qual-wan21-1088}, LR-bicubic, FreeScale, and CineScale
    blur wood texture, whereas TRACE
    resolves clearer grain and bark. For scene structure, direct inference
    and RIFLEx alter object layouts, while FreeSwim moves the cat off the
    woodpile and adds extra spoons. TRACE preserves the cat's placement
    and the spoon arrangement. These benefits also appear across other
    backbones and resolutions: the crops in Figs.~\ref{fig:qual-wan22-1440},
    \ref{fig:qual-bernini}, and~\ref{fig:qual-wan21-4k} show clearer rock edges,
    facade details, and fabric folds. At $3840\!\times\!2160$, TRACE also avoids
    the repeated woman produced by FreeSwim (Fig.~\ref{fig:qual-wan22-2160}),
    retaining the intended single subject.

    \subsection{Ablation Study}

    \noindent\textbf{Effect of CSTS.}
    CSTS raises VBench Mean from 0.7040 to 0.7518 and VQAScore from
    0.7929 to 0.9464 (Table~\ref{tab:ablation}). In Fig.~\ref{fig:ablation-visual},
    the baseline has broken leaf patterns and abrupt brightness changes.
    CSTS produces clearer tree shapes and smoother transitions from sunlit
    leaves to shade. These differences agree with the lower BRISQUE$^*$
    and NIQE$^*$ gaps, showing improved local visual fidelity.

  \end{minipage}\hfill
  \begin{minipage}[t][\dimexpr\textheight-\dp\strutbox\relax][s]{.48\textwidth}
    \vspace{0pt}
    \begingroup
    \centering
    \setlength{\abovecaptionskip}{0pt}
    \captionsetup{type=table,position=bottom}
    \caption{Ablation on a thirty-prompt subset on Wan2.1 at
      $1920\!\times\!1080$ with matched prompts and seeds. Baseline uses
      shared-canvas window attention alone; best results are bold.}%
\label{tab:ablation}\par\vspace{4pt}
    \fontsize{6.2pt}{6.8pt}\selectfont
    \setlength{\tabcolsep}{1.2pt}
    \renewcommand{\arraystretch}{0.95}
    \begin{tabular*}{\columnwidth}{@{\extracolsep{\fill}}lcccc@{}}
      \toprule
      \multirow{2}{*}{\textbf{Variant}}
      & \textbf{VBench} & \textbf{Align.}
      & \multicolumn{2}{c}{\textbf{HR Fidelity}} \\
      \cmidrule(lr){2-2}\cmidrule(lr){3-3}\cmidrule(lr){4-5}
      & Mean$\uparrow$ & VQA$\uparrow$
      & BRISQUE$^{*}\downarrow$ & NIQE$^{*}\downarrow$ \\
      \midrule
      {\fontsize{6.3pt}{6.8pt}\selectfont Baseline} & 0.7040 & 0.7929 & 0.2550 & 0.1406 \\
      {\fontsize{6.3pt}{6.8pt}\selectfont + CSTS} & 0.7518 & 0.9464 & 0.2100 & 0.1277 \\
      {\fontsize{6.3pt}{6.8pt}\selectfont + TGSC} & 0.7061 & 0.8291 & 0.2536 & 0.1398 \\
      {\fontsize{6.3pt}{6.8pt}\selectfont \textcolor{QualAccent}{\textbf{+ CSTS + TGSC}}} & \textbf{0.7528} & \textbf{0.9653} & \textbf{0.2081} & \textbf{0.1264} \\
      \bottomrule
    \end{tabular*}\par
    \endgroup\vspace{5pt}

    \begingroup
    \setlength{\fboxsep}{1.2pt}%
    \colorbox{PromptBg}{\parbox[c]{.98\linewidth}{%
        \raggedright\fontsize{7.5pt}{8.3pt}\selectfont
        \textcolor{QualAccent}{\textbf{Prompt:}}\enspace The video showcases a stunning aerial view of a modern cityscape with a mix of historic and contemporary architecture. \ldots\strut}}%
    \endgroup\par\nointerlineskip\vspace{2pt}%
    \noindent\hbox to\linewidth{\hfil
      \begin{minipage}[t]{107.32265pt}\centering
        \input{figures/visuals/comparisons/latex/bernini/HR-T2V-0015/freeswim.tex}\par\nointerlineskip\vspace{2pt}%
        \hbox to\linewidth{\hfil{\fontsize{8pt}{9pt}\selectfont\textbf{FreeSwim}}\hfil}%
      \end{minipage}\hspace{4pt}%
      \begin{minipage}[t]{107.32265pt}\centering
        \input{figures/visuals/comparisons/latex/bernini/HR-T2V-0015/ours.tex}\par\nointerlineskip\vspace{2pt}%
        \hbox to\linewidth{\hfil{\fontsize{8pt}{9pt}\selectfont\textbf{\textcolor{QualAccent}{TRACE (Ours)}}}\hfil}%
      \end{minipage}\hfil}\par
    \captionof{figure}{Bernini-R-1.3B at $1920{\times}1080$: FreeSwim and TRACE with a matched prompt and seed, with two detail crops.}%
\label{fig:qual-bernini}
    \par\nointerlineskip\vspace{6pt}

    \noindent\begin{minipage}[t]{\columnwidth}
      \centering
      {\fontsize{6.5pt}{7.2pt}\selectfont\textbf{\noindent\makebox[.25\linewidth]{Baseline}%
        \makebox[.25\linewidth]{+ TGSC}%
        \makebox[.25\linewidth]{+ CSTS}%
        \makebox[.25\linewidth]{\textcolor{QualAccent}{+ CSTS + TGSC}}}\par}%
      \nointerlineskip\vspace{1pt}
      \noindent%
      \begingroup
      \setlength{\fboxsep}{1.2pt}%
      \colorbox{PromptBg}{\parbox[c]{.98\linewidth}{%
          \raggedright\fontsize{7.5pt}{8.3pt}\selectfont
          \textcolor{QualAccent}{\textbf{Prompt:}}\enspace A cyclist wearing a yellow jacket rides along a tree-lined street \ldots\strut}}%
      \endgroup\par\nointerlineskip\vspace{1pt}
      \noindent\input{figures/visuals/comparisons/latex/ablation/DEV-TGSC-cyclist.tex}\par\nointerlineskip
      \captionsetup{type=figure,position=bottom}
      \caption{CSTS/TGSC ablation on
        Wan2.1 at $1920{\times}1080$. With CSTS alone, an extra cyclist
        appears; TGSC removes it.}%
\label{fig:ablation-visual}
    \end{minipage}\par\nointerlineskip\vspace{6pt}

    \noindent\textbf{Effect of TGSC.}
    Adding TGSC to CSTS raises VQAScore from 0.9464 to 0.9653
    (Table~\ref{tab:ablation}). Figure~\ref{fig:ablation-visual} shows the extra
    cyclist removed while the main rider and background detail remain.
    TGSC alone barely improves VBench Mean and produces patchy leaf textures.
    Together, they lead on all four metrics through scene refinement
    and local correction.

    \section{Conclusion}
    \label{sec:conclusion}
    In this paper, we propose TRACE, a training-free framework for extending
    video generation backbones to higher resolutions. Cross-scale trajectory
    synchronization (CSTS) builds early high-resolution trajectories from
    low-resolution predictions without high-resolution model calls.
    Shared-canvas window attention then refines detail, while trajectory-guided
    structural correction (TGSC) uses the low-resolution layout to remove
    extra subjects. Experiments across three backbones demonstrate improved
    perceptual quality and text alignment, with \mbox{2.0--2.9$\times$} speedups
    over FreeSwim~\citep{wu2026freeswim}. This work supports efficient high-resolution
    video generation without retraining existing models.

  \end{minipage}
\end{figure*}

\clearpage

\interlinepenalty=0
\predisplaypenalty=10000
\postdisplaypenalty=0
\setlength{\parfillskip}{0pt plus 1fil}
\bibliography{main}
\bibliographystyle{ieeenat_fullname}

\end{document}